\pdfoutput=1

\documentclass[11pt]{article}
\usepackage[final]{ACL}

\usepackage{xspace}

\newcommand{\bi}{\begin{itemize}}
\newcommand{\ei}{\end{itemize}}

\newcommand{\be}{\begin{enumerate}}
\newcommand{\ee}{\end{enumerate}}
\newcommand{\beqn}{\begin{eqnarray*}}
\newcommand{\eeqn}{\end{eqnarray*}}

\newcommand{\eg}{{\em e.g.,}\xspace}
\usepackage{multirow}

\newcommand{\kgg}{{LightKGG}\xspace}
\usepackage{booktabs}

\usepackage{mdframed}
\usepackage[most]{tcolorbox}
\newtcolorbox{bluebox}{colframe=blue, colback=white}

\title{LightKGG: Simple and Efficient Knowledge Graph Generation from Textual Data}

\author{
 \textbf{Teng Lin}
\\
 DSA,HKUST(GZ)
 \\
\\
 \small{
    \href{mailto:email@domain}{tlin280@connect.hkust-gz.edu.cn}
 }
}

\begin{document}
\maketitle

\begin{abstract}
The scarcity of high-quality knowledge graphs (KGs) remains a critical bottleneck for downstream AI applications, as existing extraction methods rely heavily on error-prone pattern-matching techniques or resource-intensive large language models (LLMs). While recent tools leverage LLMs to generate KGs, their computational demands limit accessibility for low-resource environments. Our paper introduces LightKGG, a novel framework that enables efficient KG extraction from textual data using small-scale language models (SLMs) through two key technical innovations: (1) \textbf{Context-integrated Graph extraction} integrates contextual information with nodes and edges into a unified graph structure, reducing the reliance on complex semantic processing while maintaining more key information; (2) \textbf{Topology-enhanced relationship inference} leverages the inherent topology of the extracted graph to efficiently infer relationships, enabling relationship discovery without relying on complex language understanding capabilities of LLMs. By enabling accurate KG construction with minimal hardware requirements, this work bridges the gap between automated knowledge extraction and practical deployment scenarios while introducing scientifically rigorous methods for optimizing SLM efficiency in structured NLP tasks.   


\end{abstract}
\section{Introduction}
The growing reliance on knowledge graphs (KGs) to power intelligent systems, from search engines to recommendation platforms, has exposed a critical challenge: the scarcity of high-quality, scalable methods for KG construction. Modern approaches remain bifurcated between resource-intensive paradigms, such as large language models (LLMs), which demand prohibitive computational costs, and pattern-based extraction techniques, which suffer from limited generalizability and error propagation~\cite{shenoy2021studyquality, 11107459, lin2025mebench}. While tools like KGGen~\cite{mo2025kggen} and GraphRAG~\cite{edge2024local} demonstrate the potential of LLMs for KG generation, their reliance on massive hardware infrastructure and high cost renders them inaccessible for real-world, resource-constrained scenarios. This gap between theoretical capability and practical deployability underscores an urgent need for frameworks that balance efficiency, accuracy, and accessibility.  

Our work is motivated by two fundamental observations about Small Language Models(SLMs) in graph extraction tasks: 
(1) Although small language models exhibit limitations in complex relation recognition and nuanced semantic understanding, they excel at entity identification, pattern matching, and semantic comprehension in structured contexts with clear informational boundaries~\cite{fan2025minirag,lin2025Simplifying}.
(2) Reasoning within graph structures helps mitigate the impact of noise compared with inference for text, while multi-path clues enhance error reduction, thereby reducing the performance gap between SLMs and large language models (LLMs) in knowledge extraction tasks.

In this work, we introduce LightKGG, a novel framework designed to democratize high-quality KG extraction by leveraging small-scale language models (SLMs) through two technical innovations: (1) The context-integrated graph extraction mechanism unifies entities, edges, and contextual information into a unified graph structure, bypassing the need for costly heavy semantic parsing while preserving key information. (2) The lightweight topology-enhanced inference method exploits the inherent connectivity patterns of the extracted graph to deduce relationships efficiently, circumventing the dependency on advanced linguistic reasoning typically required for LLMs. Together, these two modules enable LightKGG to construct accurate KGs from text with minimal hardware requirements, achieving performance comparable to LLM-driven baselines at a fraction of their computational cost.

\section{Related works}
The automation of knowledge graph (KG) construction from textual data has evolved through three primary paradigms: rule-based or pattern-matching systems, large language model (LLM)-driven approaches, and hybrid methods seeking to balance scalability and accuracy. 

\subsection{Rule-Based and Pattern-Matching Systems}
Early KG construction relied on handcrafted rules (\eg Hearst patterns~\cite{hearst1992automatic}) and statistical methods like OpenIE ~\cite{angeli2015leveraging}~\cite{Etzioni2008open}~\cite{Etzioni2011Open}. While lightweight, these approaches struggle with ambiguity and domain adaptation, as rigid patterns fail to capture nuanced linguistic variations~\cite{Etzioni2011Open}. Subsequent improvements, such as neural pattern learning~\cite{vashishth2018reside}, partially address these issues but still depend on annotated corpora and lack robustness to noisy text.

\subsection{LLM-Driven KG Generation} 
Recent advances in LLMs, such as GPT-4o~\cite{openai2023gpt4} and PaLM~\cite{Chowdhery2023PaLM}, have spurred frameworks like GraphRAG~\cite{edge2024local}~\cite{microsoft2024graphrag} and KGGen~\cite{mo2025kggen} that leverage these models’ semantic reasoning capabilities to extract entities and relationships. While LLMs excel at contextual understanding and generalization, their computational demands—often requiring GPU clusters for inference—render them impractical for resource-constrained environments~\cite{bommasani2022opportunitiesrisks}. Efforts to mitigate these costs, such as model distillation~\cite{hinton2015distilling}, often sacrifice relational accuracy or require extensive fine-tuning data, which remains scarce for specialized domains. 

\subsection{Hybrid and Lightweight Methods}  
Recent work explores intermediate solutions, such as coupling small language models (SLMs) with task-specific architectures~\cite{jiao2023goodtranslatoryes, lin2025srag}. For instance, KG-BERT~\cite{yao2019kgbert} fine-tunes BERT for KG completion, yet retains high memory overhead. Other efforts integrate graph neural networks (GNNs) to refine topological features~\cite{zhang2023learnable}, but their multi-layer message-passing mechanisms introduce latency incompatible with real-time applications. Crucially, these methods still prioritize semantic depth over structural efficiency, leaving the potential of topology-driven inference underexplored.  

\subsection{LightKGG}  
LightKGG diverges from prior work by rethinking how structural and semantic signals can synergize in low-resource settings. Unlike LLM-centric tools, it avoids heavy semantic parsing by unifying context and graph structure during extraction. In contrast to GNN-based systems, its topology-enhanced inference operates without iterative training, leveraging inherent graph properties (\eg node centrality, path density) for rapid relationship discovery. This dual innovation bridges a critical gap: enabling SLMs to achieve LLM-like KG fidelity while operating within stringent hardware constraints—a leap forward for democratizing knowledge-driven AI.  

By integrating insights from lightweight NLP and graph topology, LightKGG advances a scientifically rigorous yet pragmatically deployable paradigm, challenging the assumption that large-scale systems are indispensable for high-quality KG generation.
\section{LightKGG Framework}

\begin{figure*}[t!]
\begin{center}
\includegraphics[width=0.9\linewidth]{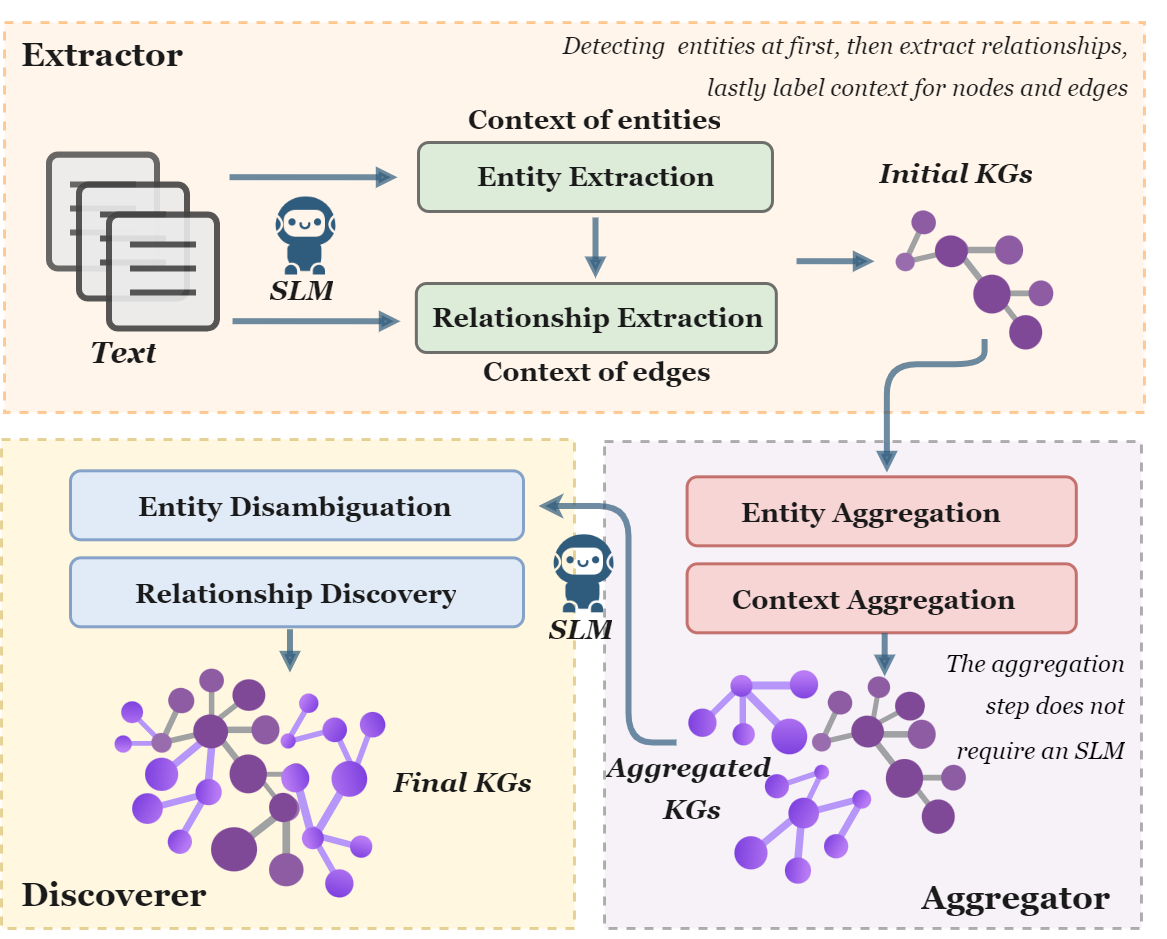}
\end{center}
\caption{Overview of LightKGG method. }
\label{fig:light}
\end{figure*}

LightKGG is designed to address the inefficiency and resource-intensity of traditional knowledge graph (KG) generation methods by leveraging small-scale language models (SLMs) in a modular, multi-stage pipeline. Unlike prior approaches that rely on large language models (LLMs) for end-to-end extraction, our framework decouples the extraction, consolidation, and discovery phases to optimize computational efficiency while maintaining semantic rigor. As illustrated in Figure~\ref{fig:light}, LightKGG comprises three core modules: \textbf{Extractor}, \textbf{Aggregator}, and \textbf{Discoverer}, each tailored to address the distinct challenges in the construction of scalable KGs.

\subsection{Context-integrated Graph Extraction}
The Extractor module employs SLMs to parse unstructured text into structured entity-relation triples while preserving contextual metadata (\eg temporal/spatial cues or entity attributions). By jointly modeling entities, relations, and their context as a unified graph substructure, this stage minimizes semantic ambiguity without resorting to LLM-scale reasoning. For example, in the sentence ``Marie Curie discovered radium in 1898'' the SLM extracts the triple \colorbox{black!8}{(Marie Curie, discovered, radium)} while annotating the relation with the temporal context \colorbox{black!8}{(year: 1898)}. This contextual anchoring enables lightweight disambiguation during downstream stages, reducing dependency on exhaustive semantic parsing.

\subsection{Graph Aggregation}
The Aggregator module synthesizes entity-relationship substructures from multiple textual sources to construct a globally consistent knowledge graph (KG). This stage centers on integrating nodes, edges, and their associated contextual metadata, with node names standardized to lowercase to ensure uniformity. For contextual harmonization, which resolves discrepancies by preserving complementary information. For example, if one source labels ``Alan Turing'' as a ``computer scientist'' and another as a ``mathematician'', the module retains both descriptors as coexisting attributes rather than discarding or conflating them, such as ``Alan Turing \{(occupation: computer scientist, mathematician)\}''. The result is a unified KG where nodes and edges are augmented with multi-context metadata, enhancing the graph’s capacity for context-driven inference.  

\subsection{Topology-enhanced Relation Discovery}
In LightKGG, we propose a lightweight topology-enhanced relation inference mechanism that leverages these structural features, such as node connection density, path length, and degree centrality, to efficiently infer latent relationships. The Discover module capitalizes on the consolidated graph’s topology to reduce reliance on computationally intensive semantic parsing. By analyzing structural patterns (\eg connection paths, node degrees), it enables small-scale language models to perform relationship reasoning with minimal linguistic depth. Key applications include:

\bi
    \item \textbf{Entity Disambiguation}: Ambiguous entities are resolved using their topological context. For instance, if ``Apple'' is densely connected to nodes like ``Company'' and ``Smartphone'', during the inference phase, the model can readily identify the entity as a company rather than a fruit.
    
    \item \textbf{Confidence Reinforcement}: Relationships supported by multiple independent paths (\eg ``Einstein → worked\_at → Princeton → collaborated\_with → Gödel'' and ``Einstein → influenced → Gödel'') are assigned higher confidence scores, mitigating influence from sparse or conflicting sources.
    
    \item \textbf{Implicit Relationship Identification}: Graph traversal algorithms (we use bidirectional BFS) and probabilistic rule mining (\eg transitive closure) uncover indirect connections of nodes. For example, a path like ``Curie → mentor → Meitner → colleague → Fermi'' may imply a latent ``scientific lineage'' relationship between Curie and Fermi.
\ei

All in all, our topology-based graph relationship discovery approach enables small-scale models to efficiently uncover implicit relationships without depending on large-scale models to do deep semantic analysis of plain text.
\section{Experiment}
\subsection{Settings}
To evaluate LightKGG’s performance, we conduct experiments for graph generation accuracy. We compare LightKGG against three baselines:
KGGen~\cite{mo2025kggen}, OpenIE~\cite{angeli2015leveraging}, GraphRAG~\cite{edge2024local}. For test datasets, we use a subset(100 sentences) of SciERC~\cite{luan2018multitask} for scientific entity/relation extraction which using F1 scores for evaluation of entity/relation extraction, and MINE~\cite{mo2025kggen}, which measures knowledge graph extractor’s ability to condense unstructured text inputs into KG with MINE-score for evaluation. For SLMs, we conduct experiments using the representative Phi-3.5-mini-instruct~\cite{abdin2024phi3}, GLM-Edge-1.5B-Chat~\cite{thudm2024glmedge} and DeepSeek-R1-Distill-Qwen-1.5B~\cite{deepseekai2025deepseekr1}. For LLM, we use  the widely recognized GPT-4o model~\cite{openai2023gpt4}. All experiments are conducted on a single NVIDIA RTX4090 GPU (24GB VRAM) to simulate resource-constrained environments. 

\subsection{Ablation Study}
To rigorously evaluate the contributions of LightKGG’s core components, we conduct an ablation study on the full framework, which employs the Phi-3.5-mini-instruct (SLM) as default configuration. This analysis isolates the impact of each design innovation by systematically disabling key features and comparing performance against the complete system. Specifically, we assess:
(1) Context integrating: Removes context of nodes and edges during entity-relation extraction.
(2) Topology-enhanced Inference: Disables Topology-based relationship discovery, relying solely on direct textual extractions.
(3) SLM vs. LLM Trade-offs: Replaces Phi-3.5-mini-instruct model with a large language model (GPT-4o) to quantify accuracy gains against computational and economic costs.

\begin{table*}[t!]
  \centering
    \caption{Experimental results for SciERC(a subset) and MINE.}
    \vspace{.2em}
  \begin{tabular}{lccc}
    \toprule
    \textbf{Configuration}  & \textbf{Entity-F1  } & \textbf{Relation-F1  } & \textbf{MINE-scores}\\
    \midrule
    KGGen(GPT-4o)  &0.891 &0.853 &0.685\\
    GraphRAG(GPT-4o)  &0.826 &0.785 &0.501\\
    OpenIE  &0.685 &0.617 &0.298\\
    \textbf{\kgg(Phi)}&\textbf{0.856} &\textbf{0.831} &\textbf{0.673}\\
    \textbf{\kgg(GLM)}&\textbf{0.832} &\textbf{0.821} &\textbf{0.638}\\
    \textbf{\kgg(DeepSeek)}&\textbf{0.788} &\textbf{0.736} &\textbf{0.567}\\
   \bottomrule
  \end{tabular}
  \label{tab:main}
\end{table*}

\begin{table*}[t!]
  \centering
    \caption{Experimental results for ablation study.}
    \vspace{.2em}
  \begin{tabular}{lccc}
    \toprule
    \textbf{Methods}  & \textbf{Entity-F1  } & \textbf{Relation-F1  } & \textbf{MINE-scores}\\
    \midrule
    \kgg(Phi)&0.856 &0.831 &0.673\\
    w/o Context-enrich  &0.798 &0.721 &0.568\\
    w/o Topology Inference  &0.856 &0.732 &0.611\\
    SLM replaced with GPT-4o  &0.885 &0.851 &0.717\\
   \bottomrule
  \end{tabular}
  \label{tab:ablation}
\end{table*}

\subsection{Main Results}

\textbf{LightKGG(Phi) vs. LLM Baselines:} As shown in Table~\ref{tab:main}, LightKGG(hi-3.5-mini-instruct) achieves 96\% of KGGen’s Entity-F1 (0.856 vs. 0.891) and 97\% of its Relation-F1 (0.831 vs. 0.853) despite using a small language model. The gap widens in MINE scores (0.673 vs. 0.685), suggesting LightKGG’s topology-enhanced inference partially compensates for SLMs’ weaker semantic understanding but still trails LLM-driven reasoning in holistic KG quality.

\textbf{LightKGG Variants:} We can see from Table~\ref{tab:main} that performance degrades with smaller SLMs: Phi > GLM > DeepSeek, indicating model architecture and scale directly impact extraction accuracy. Notably, even LightKGG (DeepSeek) outperforms GraphRAG (GPT-4o) in MINE scores (0.567 vs. 0.501), highlighting the framework’s ability to leverage graph topology to offset SLM limitations.


\subsection{Ablation study results}
As shown in table~\ref{tab:ablation}, the ablation study demonstrates that LightKGG’s context integration is the most critical component, with its removal causing severe performance declines (\eg 13.2\% drop in Relation-F1 and 15.6\% lower MINE-score), underscoring its role in disambiguating entities and inferring implicit relationships. Disabling topology-based inference significantly harms relational coverage (11.9\% Relation-F1 reduction), validating its ability to uncover latent connections via graph structure, while replacing the SLM (Phi-3.5) with GPT-4o boosts accuracy marginally (+2.4–6.5\%) but incurs high costs. Overall, LightKGG’s design optimally balances efficiency and effectiveness, prioritizing context-aware extraction and topological reasoning for scalable knowledge graph construction, with SLMs offering a practical trade-off for resource-constrained scenarios.

\subsection{Analysis of LightKGG}

For Efficiency-Accuracy trade-off, LightKGG (Phi) achieves ~85\%+ F1 scores with a fraction of LLM computational costs (\eg more than 10× smaller model size), validating its design for resource-constrained environments. For Topology-enhanced KG quality, LightKGG (Phi) narrows the MINE gap with KGGen (0.673 vs. 0.685) by exploiting graph structure (\eg path analysis, node centrality) to infer implicit relationships, reducing reliance on LLM-scale semantic depth.

LightKGG demonstrates that SLMs + graph topology can rival LLM-driven KG extraction in accuracy while being vastly more efficient. However, LLMs (KGGen) remain superior for tasks requiring deep semantic inference (\eg rare or ambiguous relationships).

\section{Limitations}
For model dependency, LightKGG’s performance varies by SLM choice (e.g., Phi vs. DeepSeek), suggesting domain-specific SLM fine-tuning is critical. In the aspect of complex relation handling, the Relation-F1 drop between LightKGG (Phi) and KGGen (0.831 vs. 0.853) reflects challenges in capturing nuanced relationships (\eg ``inhibits'' vs. ``regulates'') without LLM-level contextual reasoning.

\section{Conclusion}

In conclusion, LightKGG presents a scalable and resource-efficient framework for generating high-quality knowledge graphs (KGs) from textual data using small language models (SLMs), addressing critical limitations of existing LLM-dependent or pattern-based methods. By integrating context-integrated graph extraction and topology-enhanced relationship discovery, the framework reduces reliance on costly intensive semantic processing while maintaining accuracy. The unified graph structure captures entities, relationships, and contextual cues in a cohesive manner, enabling robust inference even with limited linguistic understanding. Experimental results demonstrate that LightKGG achieves competitive performance in KG construction while drastically lowering hardware requirements, making it accessible for low-resource environments. This advancement bridges the gap between theoretical knowledge extraction and real-world deployment, particularly in domains where efficiency and interpretability are paramount. Future work could extend this approach to multilingual and multimodal data, further optimizing SLM capabilities for structured NLP tasks. LightKGG not only democratizes KG generation but also sets a foundation for resource-conscious AI systems, emphasizing the untapped potential of SLMs in complex information extraction challenges.



\bibliography{ref/custom}

\end{document}